# Kinetostatic Analysis and Solution Classification of a Planar Tensegrity Mechanism

P. Wenger[1] and D. Chablat[1]

[1] *Laboratoire des Sciences du Numérique de Nantes, UMR CNRS 6004*
*Ecole Centrale de Nantes, 1, Rue de la Noe, 44321 Nantes, France,*
*email: [philippe.wenger, damien.chablat]@ls2n.fr*

**Abstract:** Tensegrity mechanisms have several interesting properties that make them suitable for a number of applications. Their analysis is generally challenging because the static equilibrium conditions often result in complex equations. A class of planar one-degree-of-freedom (dof) tensegrity mechanisms with three linear springs is analyzed in detail in this paper. The kinetostatic equations are derived and solved under several loading and geometric conditions. It is shown that these mechanisms exhibit up to six equilibrium configurations, of which one or two are stable. Discriminant varieties and cylindrical algebraic decomposition combined with Groebner base elimination are used to classify solutions as function of the input parameters.

## 1 Introduction

A tensegrity structure is an assembly of compressive elements (struts) and tensile elements (cable, springs) held together in equilibrium [1, 2, 3].Their inherent interesting features (low inertia, natural compliance and deployability) make them suitable in several applications. They can also be used as preliminary models in musculo-skeleton systems to analyze animal and human movements [4, 5]. A spine can be modelled by stacking a number of suitable tensegrity modules. Accordingly, the frame of this work is a preliminary step of a large collaborative project with the Museum National d'Histoire Naturelle (MNHN) to model bird necks.
A tensegrity mechanism can be obtained by actuating one or several elements. Most results on tensegrity mechanisms have been published recently, see for example [6, 7, 8] and references therein. Deriving the input/output equations of a tensegrity mechanism needs to solve the equilibrium conditions. They are generally obtained by minimizing the potential energy, which often leads to complex equations. Planar tensegrity mechanisms (PTM) are simpler to analyze and are more suitable for algebraic computations. A 2-DOF PTM was analyzed by Arsenault [6] in terms of its kinetostatics, dynamics and workspace. Recently, Boehler [8] proposed a more complete definition of the workspace of that 2-DOF PTM, along with a method with higher-order continuation tools to evaluate it. This work focuses on a one-DOF PTM made of one base telescopic rod, two crossed fixed-length rods and three connecting springs (see fig. 1). This mechanism was studied in [9] in the particular case of symmetric geometric and loading conditions. The equilibrium configurations were solved for a set of geometric parameters and for one actuator input value. Here, this class of PTM is analyzed in a more systematic way and in more details, with the goal of understanding in depth the evolution of the number of stable and unstable solutions as function of the geometric parameters, the loading conditions and the actuated joint inputs. It turns out that the algebra involved in the stability analysis may prove very complicated while the PTM at hand is rather simple. Discriminant varieties and cylindrical algebraic decomposition are used to classify the number of stable solutions as function of some input parameters. It is shown that there are always up to six equilibrium solutions, of which at most one or two are stable.

## 2  Mechanism description and basic equations

The studied mechanism is shown in fig. 1. It is composed of two rigid rods $A_1A_3$ and $A_2A_4$ of lengths $L_1$ and $L_2$ and three identical linear springs of stiffness $k$ connecting $A_1A_4$, $A_2A_3$ and $A_3A_4$, respectively. A reference frame is attached to point $A_1$ with the x-axis oriented along $A_1A_2$. Point $A_1A_2$ is fixed and $A_2$ can be translated along the x-axis by a prismatic actuator. This mechanism has three dof, one is controlled by the actuator ($\rho$) and the other two result from the compliant rotations of the two struts about $A_1$ and $A_2$ denoted by $\theta_1$ and $\theta_2$, respectively.

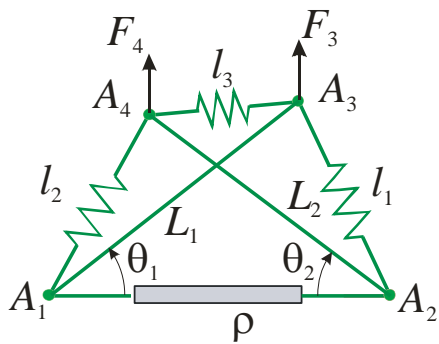

**Figure 1 Planar tensegrity mechanism**

Two vertical forces $F_3$ and $F_4$ are applied at nodes $A_3$ and $A_4$, respectively. We first consider the case of zero free length springs. This is not a purely theoretical hypothesis since equivalent zero free lengths springs can be designed as shown for example in [6] and [9]. Assuming no friction and infinite stiffness in the rods and in the prismatic joint, the potential energy $U$ of this mechanism can be written as:

$$U = \frac{k}{2}\sum_{i=1}^{3}l_i^2 - F_3 y_3 - F_4 y_4 \qquad (1)$$

where $y_3$ and $y_4$ are the ordinates of $A_3$ and $A_4$, respectively. After expressing the spring lengths $l_i$ as function of the other geometric parameters, $U$ can be shown to take the form:

$$U = \frac{k}{2}(3\rho^2 - 4L_1\rho\cos(\theta_1) + 2L_1^2 + 2L_2^2 - 4L_2\rho\cos(\theta_2) + 2L_1L_2\cos(\theta_1+\theta_2)) \\ - F_3 L_1 \rho \sin(\theta_1) - F_4 L_2 \rho \sin(\theta_2) \qquad (2)$$

The mechanism is in equilibrium when the two derivatives of $U$ with respect to $\theta_1$ and $\theta_2$ vanish simultaneously, which yields the following two equations:

$$L_1 \sin(\theta_2+\theta_1) - 2\rho\sin(\theta_2) + F_4' \cos(\theta_2) = 0 \qquad (3)$$
$$L_2 \sin(\theta_2+\theta_1) - 2\rho\sin(\theta_1) + F_4' \cos(\theta_1) = 0 \qquad (4)$$

where $F_4' = F_4/k$ and $F_3' = F_3/k$. The solutions to the direct kinetostatic problem (DKSP) of the mechanism for a given input $\rho$, are obtained by solving Eqs (3) and (4) for $\theta_1$ and $\theta_2$. Note that both stable and unstable solutions will be obtained at this stage.

## 3 Solutions to the DKSP

Equations (3) and (4) above are transformed into polynomial equations by resorting to the tan-half-angle substitutions $t_1=\tan(\theta_1/2)$ and $t_2=\tan(\theta_2/2)$ :

$$F_3' t_1^2 t_2^2 + 2L_2 t_1^2 t_2 + 2L_2 t_2^2 t_1 + 4\rho t_2^2 t_1 + F_3' t_1^2 - F_3' t_2^2 - 2L_2 t_1 - 2L_2 t_2 + 4\rho t_1 - F_3' = 0 \quad (5)$$
$$F_4' t_1^2 t_2^2 + 2L_1 t_1^2 t_2 + 2L_1 t_2^2 t_1 + 4\rho t_1^2 t_2 + F_4' t_1^2 - F_4' t_2^2 - 2L_1 t_1 - 2L_1 t_2 + 4\rho t_2 - F_4' = 0 \quad (6)$$

After elimination of one of the variables (say $t_1$) in the above two equations, a polynomial of degree 6 is obtained after clearing the factor $(1+t_1^2)$. For each root, Eqs (5) and (6) can be combined to eliminate the terms of degree 2 and $t_2$ is then solved with a linear equation. Thus, the mechanism may have up to 6 solutions to the DKSP.

It is clear that not all the solutions are stable equilibrium configurations in general. Stable solutions can be sorted out by verifying that the 2×2 Hessian matrix $H$ is definite positive, namely, if its leading principal minors are greater than zero: $H(1,1)>0$ and $\det(H)>0$.

We now inspect particular conditions that lower the degree of the above polynomial or lead to interesting special cases.

## 3.1 No external loading ($F_3=F_4=0$)

When $F_3=F_4=0$, Eqs (3) and (4) simplify to:

$$L_1 \sin(\theta_2 + \theta_1) - 2\rho\sin(\theta_2) = 0 \quad (7)$$
$$L_2 \sin(\theta_2 + \theta_1) - 2\rho\sin(\theta_1) = 0 \quad (8)$$

Thus, $\theta_i = 0$ or $\pi$, i=1,2 are solutions to the above system, which give four singular configuration (the mechanism is fully flat and cannot resist any force along the vertical direction). There are two more solutions of the form:

$$\begin{aligned}(\theta_1 = \arctan(Q/R_1), \theta_2 = \arctan(Q/R_2)), \\ (\theta_1 = -\arctan(Q/R_1), \theta_2 = -\arctan(Q/R_2))\end{aligned} \quad (9)$$

Moreover, when the coordinates of $A_3$ and $A_4$ are calculated with the above solutions, it is found that $y_3=y_4$ and $x_3 - x_4 = \rho$, which means that the mechanism remains always in a parallelogram configuration, even when $L_1 \neq L_2$.

## 3.2 Symmetric design and equal forces

When $L_1=L_2$ and $F_3=F_4$, the system is fully symmetric. This situation was studied by Arsenault [9] under the assumption that all solutions satisfy $\theta_1=\theta_2$. Accordingly, the DKSP was solved with only one equation (the derivative of U w.r.t. $\theta =\theta_1=\theta_2$), resulting in a 4$^{th}$ degree polynomial equation. In fact, it is not proven that the solutions are always of the form $\theta_1=\theta_2$ and the DKSP is solved here with $\theta_1 \neq \theta_2$ a priori. Thus, we use the two equations (7) and (8). To get simpler expressions, the second equation is subtracted to the first one. Then the tan-half substitution is done in this new equation and the following new system is obtained:

$$(t_1 - t_2)(F_4(t_1 + t_2) + 2k\rho(1 - t_1 t_2)) = 0 \quad (10)$$
$$F_4(t_1^2 t_2^2 - 1) + 2L_1 k(t_1^2 t_2 + t_1 t_2^2) + 4k\rho(t_1^2 t_2 + t_2) + F_4(t_2^2 - t_1^2) - 2L_1 k(t_1 + t_2) = 0 \quad (11)$$

The first factor $(t_1 - t_2)$ appearing in (10) confirms that solutions $\theta_1=\theta_2$ exist but the second factor indicates that solutions with distinct angles may also appear. Eliminating $t_1$ from the second factor of (10) and (11) leads to a polynomial of degree 2 in $t_2$:

$$4F_4 k^2 t_2^2 (L\rho + \rho^2) + 16k^3 \rho^3 t_2 + F_4^3(t_2^2 - 1) + 4F_4^2 k\rho t_2 + 4F_4 k^2 (L\rho - \rho^2) = 0 \quad (12)$$

Since $t_1$ can then been solved linearly using the second factor of (10), there are up to two solutions of $\theta_1 \neq \theta_2$. Moreover, the two solutions are of the form $(t_1, t_2)$ and $(t_2, t_1)$ since the same polynomial as (12) could have been obtained in $t_1$ by eliminating $t_2$ instead of $t_1$. The equal solutions obtained from the first factor in

(10) are calculated by substituting $t_2=t_1$ in eq (11), which yields up to 4 distinct solutions.

### 3.3 Stability analysis

The leading principal minors of the Hessian matrix $H$ must be calculated and their sign must be positive for an equilibrium solution to be stable. Their expression is large and is not reported here.

In the symmetric case, ($L_1=L_2$ and $F_3=F_4$), the symbolic calculation of $\det(H)$ for the solutions $\theta_1 \neq \theta_2$ is tractable. Solving the second factor of (10) for $t_2$ and replacing the solution in $\det(H)$ leads to the following expression:

$$\det(H) = -\frac{4(t_2^2+1)^2(4k^2\rho^2+F_4^2)^2(-k\rho t_2^2+F_4 t_2+k\rho)^2}{(-2k\rho t_2+F_4)^4} \quad (13)$$

which is always negative. Thus, the two equilibrium solutions satisfying $\theta_1 \neq \theta_2$ are always unstable.

For the unloaded case ($F_3=F_4=0$), it can be easily shown by reporting $\theta_{1,2} = 0$ or $\pi$ into $\det(H)$ and $H(1,1)$ that three solutions of the four flat ones are always unstable.

Regarding the other cases, symbolic calculations did not succeed and no general results could be obtained at this stage.

In the next section, the number of solutions according to the inputs parameters is investigated using more sophisticated tools, namely, cylindrical algebraic decomposition (CAD).

## 4 Solutions classification using CAD

In this section the number of stable equilibrium solutions is classified as function of the geometric and physical parameters of the PTM. The algebraic problem relies on solving a polynomial parametric system of the form:

$$E = \{\mathbf{v} \in \square^n, p_1(\mathbf{v}) = 0, \ldots, p_m(\mathbf{v}) = 0, q_1(\mathbf{v}) > 0, \ldots, q_l(\mathbf{v}) > 0\} \quad (14)$$

Such systems can be solved in several ways. Discriminant varieties (DV) [10, 12, 13] and CAD [11, 12, 13] are used here. They provide a formal decomposition of the parameter space through an algebraic variety that is known exactly. These tools have already been applied successfully in similar classes of problems [12], [13]. Roughly speaking, DV generate a set of separating hyper-surfaces in the parameters space of the parametric system at hand, such that the number of solutions in each resulting connected components or cells is known and constant. The DV can be computed with known tools like Groebner bases using the Maple sub-package *RootFinding[parametric]*. Once the DV are obtained, an open CAD is computed to provide a description of all the cells. The number of solutions in each cell is determined by solving the polynomial system for one arbitrary point in each cell. Finally, adjacent cells with the same number of solutions are merged.

The equilibrium solutions depend on three geometric parameters (the rod lengths $L_1$ and $L_2$ and the input variable $\rho$) and two physical parameters (the spring stiffness k and the forces $F_3$ and $F_4$). However, the lengths parameters can be normalized with $L_1$ and $F_3$ and $F_4$ can be replaced by $F_3/k$ and $F_4/k$, without loss of generality. Finally our system depends on 4 independent parameters only. In what follows, $L_1$ and $k$ are fixed to 1 and 100, respectively.

## *4.1 No external loading*

We starts with the simplest situation were $F_3=F_4=0$.

We were able to show in the preceding section that three of the four flat solutions were unstable but no general information regarding the two non-flat solutions could be obtained. Since $F_3=F_4=0$, the parameter space is a plane ($L_2$, $\rho$). Computing the DV and the CAD for this case leads to the existence of a region in the parameter plane where the PTM has two stable solutions. Outside this region, the PTM has one stable solution. Figure 2 (left) shows a representation of the CAD for $L_2$ and $\rho$ in [0, 4]. The 2-solution region is the red one. The DV that bound the regions are defined by $2\rho-L_2-1=0$, $2\rho-L_2+1=0$ and $2\rho+L_2-1=0$. Here it can be easily verified with geometric arguments that these boundaries correspond to the fully flat (singular) configurations of the PTM. In the 1-solution regions, the PTM has one stable fully flat solution and in the 2-solution regions, it has two stable (non-flat) solutions, one being the mirrored image of the other as shown in fig 2 (right). The 2-solution region is of constant width equal to 1 (in fact $L_1$) when $L_2>1$, while it decreases linearly with $L_2$ when $L_2<1$.

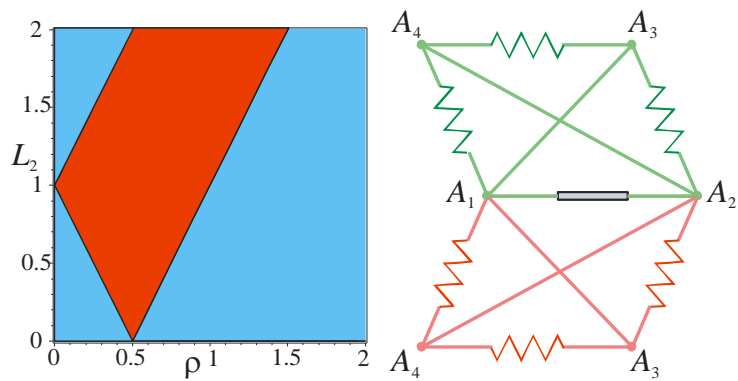

**Fig. 2:** Unloading case: CAD (left) and stable solutions for $\rho=1$, $L_2=3/2$ (right)

## *4.2 Fully symmetric case*

We now study the case $L_1=L_2$ and $F_3=F_4$. We could show in the preceding section that the two solutions $\theta_1 \neq \theta_2$ are always unstable but we could not conclude for the four solutions $\theta_1 = \theta_2$. Here the parameter space is the plane ($\rho$, $F_4$). The computed DV and CAD is illustrated in fig. 3 (left) for $0<\rho\leq 2$ and $-10\leq F_4\leq 0$ (a symmetric pattern is obtained for $0\leq F_4\leq 10$).

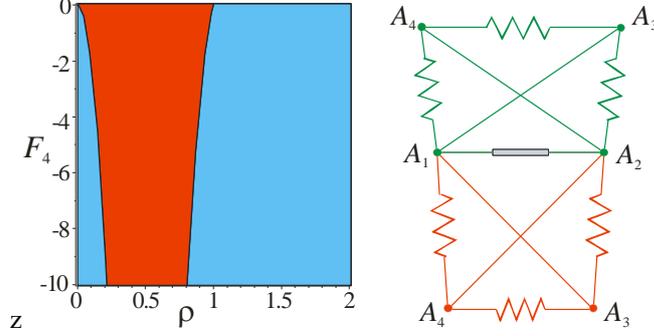

**Fig. 3** Symmetric case: CAD (left) and stable solutions for $\rho=3/4$, $F_4=-10$ (right)

It reveals that there exists a region with two stable solutions in the parameter plane. Fig 3 (right) shows two stable solutions for $\rho=3/4$ and $F_4=-10$.

The boundaries here are two curves of degree 6 defined by:

$$F_4^6 + 12\times 10^4 F_4^4 \rho^2 + 48\times 10^8 F_4^2 \rho^4 + 64\times 10^{12} \rho^6 - 16\times 10^8 F_4^2 \rho^2 = 0 \quad (15)$$

$$F_4^6 + 12\times 10^4 F_4^4 \rho^2 + 48\times 10^8 F_4^2 \rho^4 + 64\times 10^{12} \rho^6 - 12\times 10^4 F_4^4 + 336\times 10^8 F_4^2 \rho^2 \\ -192\times 10^{12} \rho^4 + 48\times 10^8 F_4^2 + 192\times 10^{12} \rho^2 - 64\times 10^{12} = 0 \quad (16)$$

In the 1-solution region, it can be shown that the PTM operates always in a reverse configuration, namely, $y_3$ and $y_4$ are negative. Assuming that the mechanism starts from a configuration with $y_3$ and $y_4$ positive, the operation range for a given $F_4$ is thus determined by the 2-solution region. The operation range decreases when the external force increases (in magnitude). It can be verified that for $F_4=0$, the operation range reaches its maximal value, which is equal to 1 (or $L_1$) in accordance with the preceding result. Note that in the presence of pulling forces ($F_4>0$), the operation range of the PTM would be full because in this case $y_3$ and $y_4$ turn out to be positive in the one-solution region.

### *4.3 General case*

The parameter space is now defined by ($\rho$, $L_2$, $F_3$, $F_4$). Two parameters are first assigned in order to have a parameter space of dimension 2. Accordingly, the DV and the CAD are computed for $F_3=F_4=-10$. Figure 4 (left) shows the obtained partition of the parameter plane ($\rho$, $L_2$) for $0<\rho\leq 2$ and $0\leq L_2\leq 2$. It looks similar to the unloaded case but the boundaries here are three curves of degree 12 in $\rho$ and in $L_2$. Their equations contain hundreds of terms. There are two stable solutions in the red region and only one in the blue regions. Figure 4 (right) shows two stable solutions for $L_2=3/2$ and $\rho=7/10$.

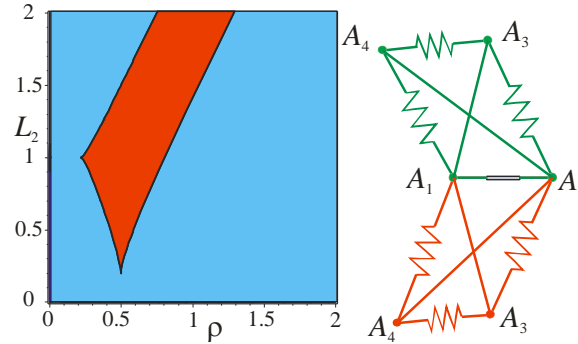

**Fig. 4** General case with $F_3=F_4=-10$: CAD (left) and stable solutions for $L_2=3/2$ and $\rho=7/10$ (right)

Like in the preceding case, the operation range is determined by the 2-solution region if the PTM starts with $y_3>0$ and $y_4>0$. The operation range reaches its maximal width for $L_2=1$, which is the fully symmetric case (it can be verified that this range is exactly the same as the one calculated from the DV above for $F_4=-10$). The operation range decreases slowly when $L_2$ increases from 1 but the decrease is much more significant when $L_2$ decreases from 1.

We now compute the DV and the CAD with $F_3=-10$ and $L_2=3/2$ in the parameter plane $(\rho, F_4)$. The result is shown in fig. 5 for $0<\rho\leq 2$ and $-30\leq F_4\leq 0$, where the red region contains 2 stable solutions and the blue region only 1. The boundaries are defined by curves of degree 12 in $\rho^2$ and in $F_4$.

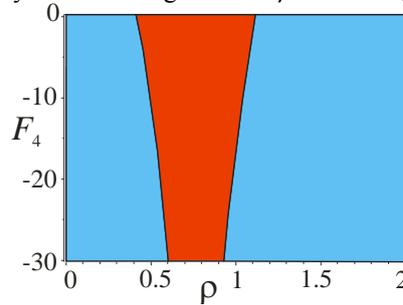

**Fig. 5** CAD for the general case with $F_3=-30$ and $L_2=1$

## 5 Discussion on more general cases

### 5.1 Adding horizontal force components

When horizontal force components $F_{3x}$ and $F_{4x}$ are added, it can be shown that this does not change the global nature of the algebraic equations and of the results. Indeed the only changes are the additional term $2F_{3x}$ (resp. $2F_{4x}$) appearing in the coefficients of $t_2^2 t_1$ and of $t_1$ in Eq. (5) (resp. of $t_1^2 t_2$ and of $t_2$ in Eq. (6)). Globally one comes up with a system yielding 6 solutions, of which 1 or 2 are stable like above.

## 5.2 Springs with non-zero free lengths

Zero free length springs have been assumed so far. It is interesting to investigate the changes induced on the algebraic complexity of the systems when non-zero free lengths are introduced in the springs. Crane et al. reported an amazing increase in complexity for a planar pre-stressed parallel manipulator made of a triangular base and platform connected by one extensible RPR leg and two springs [7]. When a free length $l_0$ is introduced in all springs, Eq. (1) becomes:

$$U = \frac{k}{2}\sum_{i=1}^{3}(l_i - l_0)^2 - F_3 y_3 - F_4 y_4 \qquad (17)$$

The point is that $l_i$ is calculated using a square root, which disappears if $l_0=0$ but remains when $l_0 \neq 0$. The two derivatives of $U$ now contain several square roots, which can be cleared after squaring two times. As a result, equations (5) and (6) become extremely complex and their degree is now 28. Due to the complex algebra, parameters must be specified before proceeding to the elimination. We could keep $\rho$ as such and all remaining parameters were assigned arbitrary values. The univariate polynomial after elimination of $t_1$ turned out to be of degree 328 in $t_2$ but could be divided by $(1+t_2^2)^{56}$, thus reducing the degree to 272. Attempts to solve the polynomial for some values of $\rho$ resulted in 12 to 30 real solutions. After verification of the vanishing of the two derivatives of $U$ to eliminate all spurious solutions, no more than 6 solutions remained. Note that for a solution to be acceptable finally, the lengths of all springs must be verified to be greater than $l_0$ in addition to the stability condition. Deeper investigations will be the subject of future work to obtain more results but it is clear that a classification study as in the case $l_0=0$ will be difficult because of high calculation times.

## Conclusions

The goal of this paper was to investigate in depth the direct kinetostatic solutions of a family of planar tensegrity mechanisms composed of a prismatic base, two crossed rods and three springs. With zero-free length springs, the problem can be treated using computer algebra tools like for the direct kinematics of parallel manipulator. We have used discriminant varieties and cylindrical algebraic decomposition to study the evolution of the number of solutions as function of the input parameters. Basically, a univariate polynomial of degree 6 must be solved in the general case, resulting in one to two stable solutions. In the unloaded case, there are always two stable symmetric solutions for a range of the input prismatic joint which is of constant width and whose limits vary with the rod lengths. Moreover, the mechanism remains always in a parallelogram configuration even when the two rod lengths are different. The mechanism can also reach one flat stable solution, which is singular. Such a stable flat solution might be of interest to store the mechanism when it is not used. When the two external forces and the two rod lengths are equal, there are still 6 solutions, including 4 unstable, non-symmetric solutions. The case of springs with non-zero free lengths was discussed and shown to lead to very large equations with high degree but no more than 6 solutions were found in our numerical experiments.

**Acknowledgments** This work is partially funded by the French ANR project "AVINECK: an arm for the bird", ANR-16-CE33-0025-02, 2017-2020.